%% file: main.tex
\documentclass{article}

\usepackage{PRIMEarxiv}

\usepackage[utf8]{inputenc} % allow utf-8 input
\usepackage[T1]{fontenc}    % use 8-bit T1 fonts
\usepackage{hyperref}       % hyperlinks
\usepackage{url}            % simple URL typesetting
\usepackage{booktabs}       % professional-quality tables
\usepackage{amsfonts}       % blackboard math symbols
\usepackage{nicefrac}       % compact symbols for 1/2, etc.
\usepackage{microtype}      % microtypography
\usepackage{lipsum}
\usepackage{authblk}

\usepackage{fancyhdr}       % header
\usepackage{graphicx}       % graphics
\graphicspath{{media/}}     % organize your images and other figures under media/ folder

\usepackage{multirow}
\usepackage{authblk}

\title{Marchuk\thanks{\textit{Gury Marchuk was a Soviet mathematician who significantly advanced weather forecasting by creating mathematical models of the atmosphere and ocean and developing numerical algorithms for prediction. His key innovation was the method of splitting (fractional steps), which made solving complex weather equations more efficient.}} : Efficient Global Weather Forecasting from Mid-Range to Sub-Seasonal Scales via Flow Matching
}

\author[1,2, $\dagger$]{Arsen Kuzhamuratov}
\author[1,2]{Mikhail Zhirnov}
\author[1,2]{Andrey Kuznetsov}
\author[1]{Ivan Oseledets}
\author[1,2]{Konstantin Sobolev}

\affil[1]{AXXX}
\affil[2]{FusionBrain Lab}
\affil[$\dagger$]{kuzhamuratov@fusionbrainlab.com}

\begin{document}
\maketitle

\begin{abstract}
Accurate subseasonal weather forecasting remains a major challenge due to the inherently chaotic nature of the atmosphere, which limits the predictive skill of conventional models beyond the mid-range horizon (approximately 15 days). In this work, we present \textit{Marchuk}, a generative latent flow-matching model for global weather forecasting spanning mid-range to subseasonal timescales, with prediction horizons of up to 30 days. Marchuk conditions on current-day weather maps and autoregressively predicts subsequent days' weather maps within the learned latent space. We replace rotary positional encodings (RoPE) with trainable positional embeddings and extend the temporal context window, which together enhance the model's ability to represent and propagate long-range temporal dependencies during latent forecasting. Marchuk offers two key advantages: high computational efficiency and strong predictive performance. Despite its compact architecture of only 276 million parameters, the model achieves performance comparable to LaDCast, a substantially larger model with 1.6 billion parameters, while operating at significantly higher inference speeds. We open-source our inference code and model at: \url{https://v-gen-ai.github.io/Marchuk/}
% Accurate global weather forecasting at subseasonal lead times remains difficult because atmospheric chaos rapidly amplifies errors in the initial state, causing forecast skill to approach climatology beyond roughly two weeks. We present \textit{Marchuk}, a latent flow-matching model for probabilistic global weather forecasting with horizons up to 30 days. \textit{Marchuk} operates in the compressed latent space of the pretrained LaDCast DC-AE, conditions on one day of 6-hourly weather maps, and generates future states autoregressively in multi-day chunks. Relative to LaDCast, \textit{Marchuk} uses learnable 2-D spatial positional embeddings together with 1-D temporal RoPE, a longer conditioning context, and variable-horizon training. In WeatherBench-2 evaluations, the 276M-parameter model improves over LaDCast-375M and remains competitive with LaDCast-1.6B while reducing 30-day ensemble inference time to 7.5 minutes on a single H100 GPU. We release code and checkpoints at: \url{https://github.com/kuzhamuratov/Marchuk}
\end{abstract}

% keywords can be removed

\input{sections/1_introduction}
\input{sections/2_related_work}
\input{sections/3_method}
\input{sections/4_experiments}

\clearpage
\input{sections/5_conclusion}
\clearpage
%Bibliography
\appendix

\input{sections/6_determine}
\input{sections/7_probabalistic}
\clearpage
\bibliographystyle{unsrt}  
\bibliography{references}

\end{document}

%% file: sections/1_introduction.tex
\section{Introduction}
\label{sec:introduction}

Subseasonal-to-seasonal (S2S) weather forecasting, encompassing the 2-to-6-week timeframe, is of critical importance to a wide range of economic sectors and societal activities. The ability to provide early warnings of anomalous conditions and support long-range planning hinges on the accuracy of predictions at these extended lead times. However, this temporal regime remains a formidable challenge for numerical weather prediction. The predictive skill of most operational models degrades rapidly beyond two weeks, often failing to outperform simple climatological benchmarks -- i.e., the average values of the past decades.
% This barrier is primarily due to the chaotic nature of the atmosphere; the stochastic evolution of its internal state, combined with unavoidable errors in initial conditions, leads to a catastrophic accumulation of error that fundamentally limits deterministic predictability \cite{14_lorenz, 3_nvidia_healda}.
This predictability barrier is fundamentally driven by the chaotic nature of the atmosphere. As highlighted by recent studies \cite{3_nvidia_healda}, unavoidable errors in initial conditions grow approximately exponentially over time-governed by the system's leading Lyapunov exponents -- leading to a catastrophic accumulation of error that fundamentally limits deterministic predictability.

Contemporary deep learning models have revolutionized mid-term forecasting. Models such as GenCast \cite{4_gencast}, WeatherNext2 \cite{5_weathernext2}, and Aurora \cite{6_aurora} have demonstrated state-of-the-art performance for lead times up to 15 days, often surpassing traditional physics-based methods. While these models represent a significant leap forward, their skill is approaching a point of saturation within this 15-day window. Furthermore, their primary challenge remains surpassing the climatology baseline at longer horizons, and this performance comes at the cost of substantial computational resources for both training and inference.

In response to the demand for longer-range forecasts, specialized S2S prediction systems are being developed. Models like the one from the Subseasonal-to-Seasonal (S2S) project - FuXi-S2S \cite{2_fuxis2s} are designed to operate on horizons extending well beyond two weeks. FuXi-S2S, for instance, simplifies the forecasting task by predicting daily mean values rather than high-frequency (e.g., 6-hourly or hourly) states, which is a pragmatic approach for extended ranges. Concurrently, generative modeling has emerged as a powerful paradigm for weather forecasting. LaDCast \cite{1_ladcast} introduced a latent diffusion framework, drawing a direct parallel to video and image generation models. By operating in the compressed latent space of a DC-AE (Deep Compressed AutoEncoder) \cite{11_dce_ae}, LaDCast demonstrated that high-quality forecasts could be generated with reduced computational demands.

Despite recent advances, important challenges remain. The current state-of-the-art latent diffusion model for weather forecasting, LaDCast, continues to exhibit degraded accuracy at extended lead times and is constrained in operational settings by its large parameter size and slow inference throughput. These limitations reduce its practical utility for extended-range probabilistic forecasting and impede deployment in time-sensitive or resource-constrained environments.

To address these gaps, we introduce Marchuk, a novel latent flow-matching Diffusion Transformer (DiT) architected for efficient and accurate forecasting from short-range through subseasonal horizons. Marchuk replaces the geometry-aware rotary positional encodings (e.g., GeoRoPE) used in LaDCast with a temporal RoPE-1D complemented by trainable spatial positional embeddings, and substantially extends the conditional context window to better capture long-range temporal variability. We further adopt a variable-length training strategy across prediction sequences, and validate our design choices through an extensive set of ablation experiments. By combining an optimized architecture with the computational advantages of latent diffusion, Marchuk attains competitive or superior predictive performance while dramatically reducing computational cost and improving inference speed, thereby making extended-range probabilistic forecasting substantially more practical.

% Finally, in the interest of reproducibility and transparency, we have made our implementation and checkpoints publicly available. We hope that Marchuk’s demonstrated efficiency and skill will encourage further research in latent-space weather forecasting.

%% file: sections/2_related_work.tex
\section{Related work}
\label{sec:related_work}

\begin{figure}[t]
    \centering
    \includegraphics[width=1.0\linewidth]{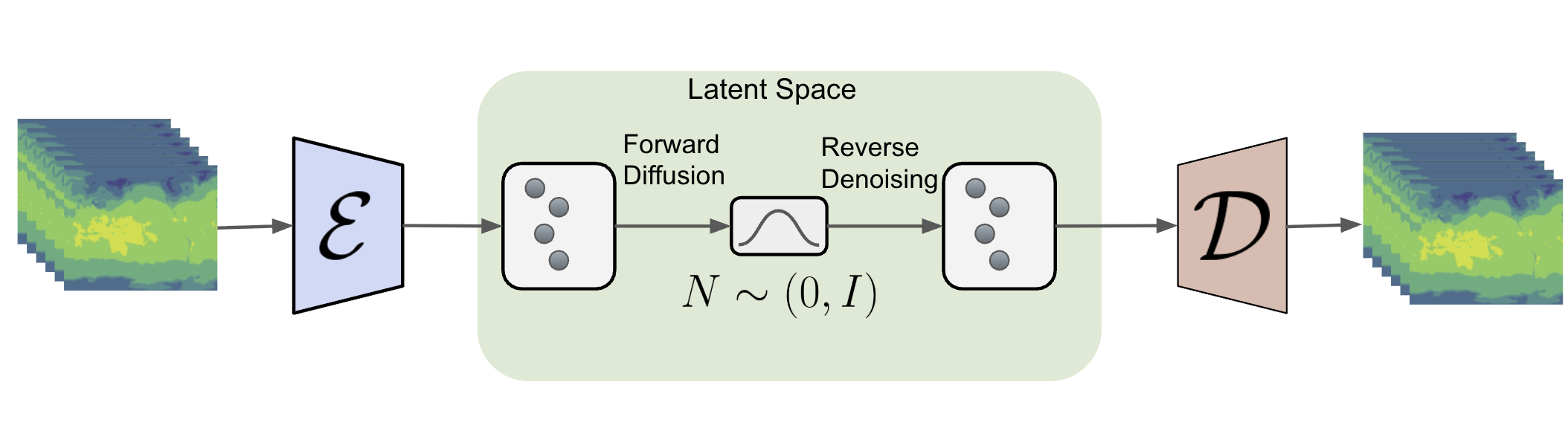}
    \caption{Marchuk operates in the latent space learned by the DC-AE model introduced in LaDCast. Within this latent space, we train a flow-matching Diffusion Transformer (DiT) to model the conditional distribution of future weather fields given the current state, enabling efficient and accurate probabilistic forecasting.}
    \label{fig:latentspace}
\end{figure}
Recent progress in machine learning for weather forecasting has produced a variety of data-driven models that significantly improve short- and medium-range prediction skill. Most state-of-the-art mid-term global forecasting models (up to ~10–15 days) operate directly on pixel-level atmospheric fields. These models are typically trained on large reanalysis datasets such as ERA5 (i.e., the 5th generation of ECMWF reanalysis
data) \cite{12_era5} and adopt architectures inspired by computer vision. For instance, GraphCast \cite{8_graphcast} employs graph neural networks (GNNs) to better represent the spherical structure of the Earth and capture long-range spatial dependencies. Other prominent approaches, including Pangu-Weather \cite{7_penguweather}, FuXi \cite{9_fuxi}, and Aurora \cite{6_aurora} rely on convolutional neural networks or transformer-based vision architectures (e.g., DiT-style blocks) to process global weather maps.

Despite their success, pixel-space models remain computationally expensive due to the extremely high dimensionality of atmospheric data. To address this limitation, recent work has explored latent-space forecasting, where atmospheric fields are first compressed before prediction. LaDCast \cite{1_ladcast} represents a key development in this direction. It introduces the first multi-modal diffusion transformer (MM-DiT) based latent model for global weather prediction. The approach employs a deep compression autoencoder (DC-AE) that compresses spatial weather fields by up to 64×, after which a transformer-based model performs the temporal forecasting in the learned latent space. Operating in latent space substantially reduces memory and computational requirements while maintaining high forecast quality. Empirical results show that latent approaches can outperform several pixel-space baselines, including Pangu-Weather, GenCast \cite{4_gencast}, and even traditional numerical weather prediction systems such as IFS-ENS for medium-range forecasts.

Evaluation of machine-learning weather models is typically performed using the WeatherBench family of benchmarks \cite{10_weatherbench2} (WeatherBench, WeatherBench-2, and WeatherBench-X). These frameworks provide standardized datasets and metrics for evaluating deterministic and probabilistic forecasting performance across multiple atmospheric variables, and they have become the de facto standard for benchmarking global weather prediction models.

Extending machine learning methods to subseasonal forecasting (2-6 week lead times) presents additional challenges. This regime is often referred to as the predictability desert, where forecast skill arises from both atmospheric initial conditions and slowly evolving boundary conditions, but neither source alone provides sufficient predictive signal. Traditional numerical weather prediction (NWP) systems address this uncertainty through ensemble forecasting; however, operational ensembles are typically limited to 4-51 members, whereas theoretical studies suggest that 100-200 members would be required to adequately represent uncertainty. Furthermore, generating dynamically balanced perturbations of the initial state remains a difficult problem.

Recent machine learning models have begun addressing these limitations. FuXi-S2S \cite{2_fuxis2s} extends earlier FuXi architectures to subseasonal forecasting by incorporating a richer set of atmospheric variables (including multiple pressure levels and surface variables) and introducing a VAE-inspired perturbation module that learns flow-dependent perturbations directly in latent space. This approach improves both deterministic and probabilistic forecast skill relative to operational NWP systems and demonstrates enhanced capability in predicting large-scale climate phenomena such as the Madden-Julian Oscillation (MJO).

Parallel advances have been made in probabilistic forecasting using diffusion models. GenCast showed that diffusion-based approaches operating directly in physical space can outperform traditional ensemble prediction systems in probabilistic metrics, albeit at high computational cost. Building on this idea, LaDCast introduces the first global latent diffusion framework for medium-range ensemble forecasting, combining a deep compression autoencoder with a transformer-based diffusion model. Additional architectural innovations -- including Geometric Rotary Position Embedding (GeoRoPE) to encode spherical geometry and dual-stream attention for efficient conditioning -- enable the model to generate hourly ensemble forecasts up to 15 days ahead. Notably, LaDCast demonstrates strong performance in predicting extreme events, producing accurate ensemble trajectories for tropical cyclones outside the training distribution.

Overall, these developments highlight two emerging directions in machine learning weather prediction: pixel-space architectures inspired by vision models and latent-space generative models that significantly reduce computational complexity while enabling probabilistic forecasting.

%% file: sections/3_method.tex
\section{Methods}

\begin{figure}[t]
    \centering
    \includegraphics[width=0.8\linewidth]{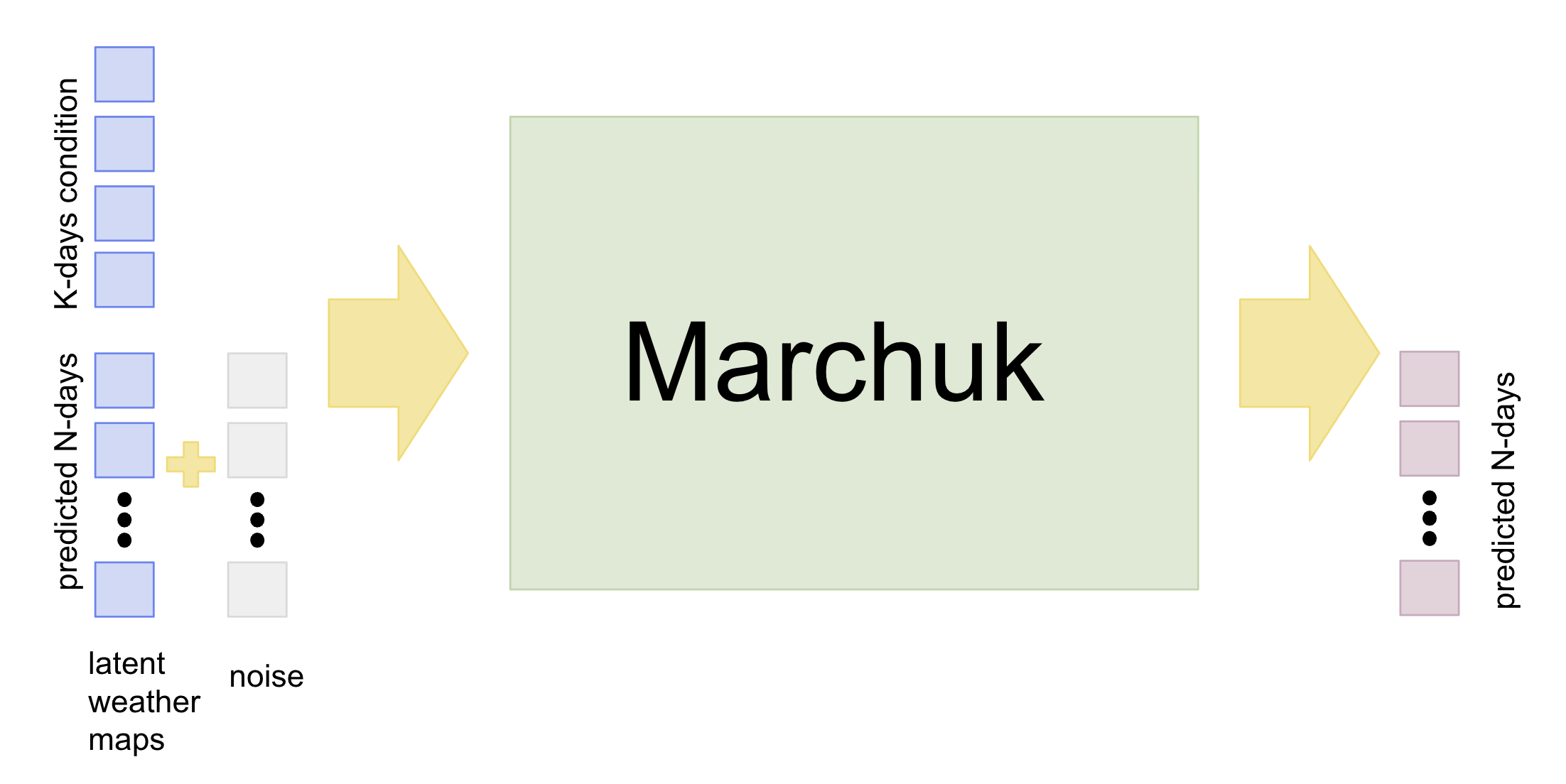}
    \caption{Marchuk is conditioned on weather maps from the previous $K$ days and is provided with noised weather maps for the subsequent $N$ days as input. The model generates refined forecasts for the next $N$ days, effectively denoising and predicting future weather states.}
    \label{fig:model_inputs}
\end{figure}
\begin{figure}[t]
    \centering
    \includegraphics[width=1.0\linewidth]{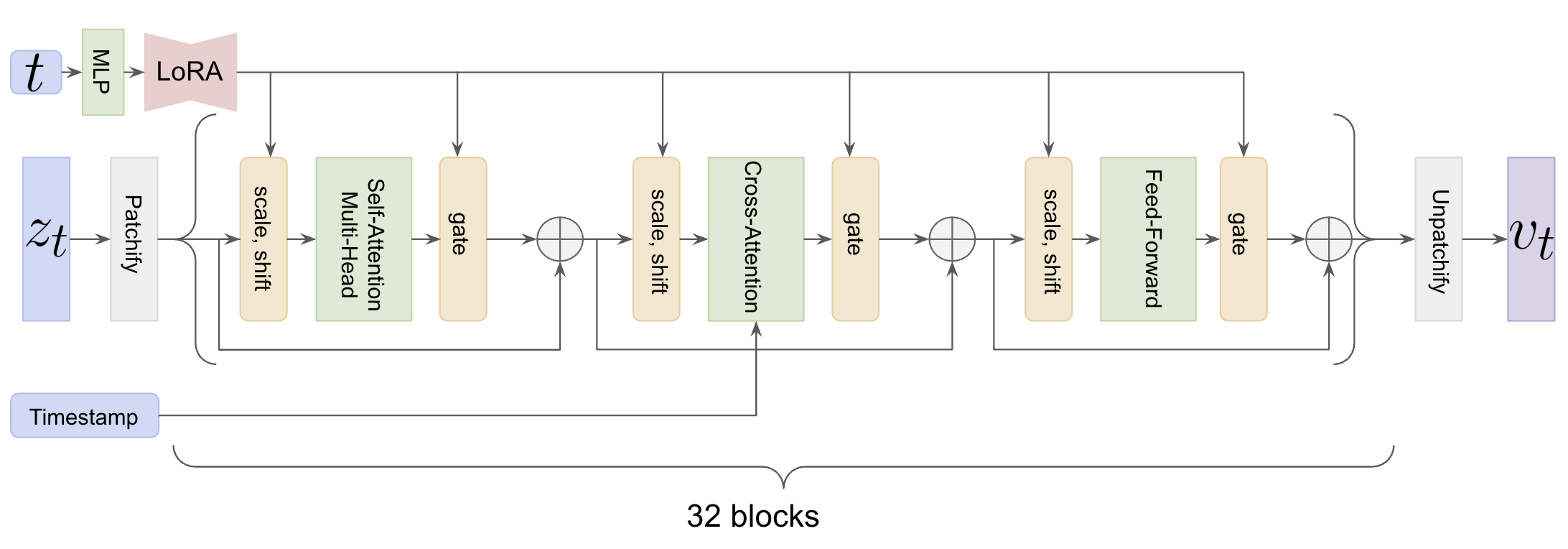}
    \caption{Marchuk is flow matching DiT model with additional architecture optimizations.}
    \label{fig:dit}
\end{figure}

\textbf{Overview.} 
We present \textit{Marchuk}, a latent generative weather-forecasting model developed within the LaDCast paradigm (see Figure~\ref{fig:latentspace}). 
Marchuk comprises two principal components. First, a \textbf{2D deep-compression autoencoder (DC-AE)} maps high-dimensional atmospheric fields into a compact latent space with approximately $64\times$ spatial compression; for our experiments we employ the pretrained DC-AE from LaDCast. 
Second, a \textbf{Diffusion Transformer (DiT)} -- trained in the DC-AE latent space -- models the temporal dynamics of these latent representations. 
Forecast generation is implemented using a Flow Matching (FM) formulation, which allows effective deterministic sampling using a relatively small number of inference steps. 
A batched forecasting strategy (producing four-day forecasts per forward pass) combined with autoregressive rollout supports ensemble prediction horizons of up to 30 days. 
Below we provide details of the dataset, model architecture, training procedure, inference strategy, and evaluation protocol.

\subsection{Data and Preprocessing}

We use the \textbf{ERA5 reanalysis dataset} from ECMWF, provided in the WeatherBench2 format. 
The dataset contains global atmospheric fields on a $1.5^{\circ}$ grid ($240\times121$ spatial resolution) with hourly temporal resolution.

Following LaDCast \cite{1_ladcast}, the input variable set includes 13 pressure levels for six atmospheric variables (geopotential, temperature, zonal wind, meridional wind, specific humidity, and vertical velocity), as well as six surface variables: 2-m temperature, 10-m zonal and meridional wind, mean sea level pressure, sea surface temperature, and 6-hour accumulated precipitation. 
This results in a total of \textbf{84 input channels}.

WeatherBench2 provides a cloud-hosted ERA5 dataset covering the period 1979-2022. 
For training we use the period \textbf{1979-2017}, while \textbf{2018-2021} is reserved for evaluation. 
Additionally, a case study from January 2026 is used for qualitative validation.

During training, a random temporal anchor is sampled at hourly resolution. 
From this anchor, a sequence of 20 time steps is extracted at 6-hour intervals (1 day of context and 4 days of forecast horizon). 
Global atmospheric fields are represented as 2D tensors obtained by unfolding the spherical grid. 
Static fields such as land-sea mask and orography are incorporated as additional channels and processed by the autoencoder.

\subsection{Model Architecture}

\paragraph{Latent Diffusion Transformer (DiT).}
Latent representations are processed by a transformer-based diffusion model that predicts their temporal evolution. The model takes as input $K$ days of conditioning information (i.e., the current atmospheric state) together with a noisy version of the subsequent $N$ days of weather states (Figure~\ref{fig:model_inputs}). The model then outputs a refined estimate of the future $N$-day weather trajectory. In our experiments, we explore different configurations of $K$ and $N$ to optimize predictive performance while reducing the number of autoregressive steps.

We adopt the Cross-DiT architecture~\cite{13_walt} and further optimize the MLP-based modulation layers following the same work. In particular, the authors propose a more parameter-efficient design for diffusion timestep conditioning by replacing per-block MLP modulation with a shared MLP layer across all DiT blocks, complemented by lightweight LoRA adaptations specific to each block. This approach significantly reduces the number of parameters associated with timestep modulation -- from approximately 40\% to less than 10\%, depending on the chosen LoRA configuration -- while allowing the model to allocate more capacity to the attention layers. A schematic overview of the architecture is provided in Figure~\ref{fig:dit}.

Another key architectural component is the incorporation of timestamp embeddings within the Cross-Attention blocks of the DiT architecture (Figure~\ref{fig:dit}). We represent timestamp in the format \textit{month:day:hour} and assign a distinct trainable embedding to each possible timestamp (366 days $\times$ 24 hours, yielding exactly 8,784 distinct embeddings). These embeddings are concatenated with the input spatial-temporal latent tokens prior to the Cross-Attention operation. This design enables the model to capture long-range temporal dependencies at the annual scale, while the joint representation with visual tokens allows the attention mechanism to selectively focus on temporally relevant features.

% The model is trained within the flow-matching framework by minimizing the mean squared error (MSE) between predicted and target latent variables under noise perturbations. To account for the non-uniform spatial distribution of grid points on the sphere, we employ a latitude-aware weighting scheme in the MSE loss, which better reflects the underlying geometry of the Earth’s surface.

\paragraph{Positional and Temporal Encoding.}
Because the latent representation preserves global spatial geometry, positional encoding must take the spherical structure of the Earth into account. LaDCast addresses this with \textit{GeoRoPE}, a geometric rotary positional embedding that decomposes spatial encoding into latitude and longitude components and extends the scheme to the temporal axis; the original authors select rotary frequencies using domain knowledge.

In Marchuk, we adopt a simplified strategy consisting of learnable 2-D spatial positional embeddings together with 1-D rotary positional embeddings (RoPE) applied along the temporal dimension. As Marchuk operates at a fixed spatial resolution, the principal advantage of 2-D RoPE -- its ability to generalize across resolutions -- is not required; consequently, a fixed set of trainable spatial embeddings provides sufficient representational flexibility while reducing the strength of hard-coded geometric priors.

\subsection{Training Procedure: Variable Horizon Training}

% During training, we randomize the forecast horizon to improve the model's robustness across multiple temporal scales.
During training, we employ a Variable Horizon Training (VHT) strategy to improve the model's robustness across multiple temporal scales. The input context is fixed at 24\,hours (four time steps at 6-hour intervals), while the prediction horizon for each training example is sampled uniformly between 1 and 8\,days. This variable horizon strategy -- in which target lengths vary across batches -- was found to yield better predictive performance and more stable autoregressive rollouts than training with a single fixed window or applying post hoc fine-tuning.

\subsection{Inference and Ensemble Forecasting}

\begin{table}[h]
\centering
\caption{\textbf{Inference Speed.} The model performs predictions entirely in the latent space for a 30-day forecast horizon with an ensemble size of 50, executed on an H100 GPU. Reported measurements account only for the forecasting component, excluding encoding / decoding.}
\label{tab:inference-speed}
\begin{tabular}{lccc}
\toprule
\textbf{Model} & \textbf{\# params}& \textbf{\#diffusion steps} & \textbf{Speed, min}\\
\hline
LaDCast & 375M &10&11\\
% \hline
LaDCast & 1.6B &10&45\\
% \hline
Marchuk & 276M&20&\textbf{7.5}\\
\bottomrule
\end{tabular}
\end{table}
During inference, forecasts are produced autoregressively in the latent space. For each forward pass the model predicts latent states for the next four days (16 time steps at 6-hour intervals); these predicted latents are then appended to the conditioning context and used as input for the subsequent prediction step, permitting rollouts to horizons of up to 30 days. Probabilistic (ensemble) forecasts are obtained by repeating the generation process with independent noise seeds: in our experiments we typically employ ensembles of 20-50 members. Sampling is carried out with the flow-matching sampler using approximately 20 integration (denoising) steps per trajectory. Operating in a highly compressed latent space substantially reduces memory and computational costs and enables efficient parallel evaluation of many trajectories. For example, Marchuk produces a 30-day ensemble forecast of 50 members in roughly 7.5 minutes on a single NVIDIA H100 GPU (see Table~\ref{tab:inference-speed}).

%% file: sections/4_experiments.tex
\section{Experiments}
\label{sec:experiments}

\subsection{Evaluation Metrics}

Model performance is evaluated using the \textbf{WeatherBench-2} \cite{10_weatherbench2} evaluation framework. 
We report both deterministic and probabilistic forecast metrics.

Deterministic metrics include:
\begin{itemize}

\item Latitude-weighted Root Mean Squared Error (RMSE)
\item Anomaly Correlation Coefficient (ACC)
\end{itemize}

Probabilistic metrics include:
\begin{itemize}
\item Continuous Ranked Probability Score (CRPS)
\item CRPS Skill
\item CRPS Spread
\end{itemize}

Metrics are averaged over latitude and time and evaluated on standard meteorological variables such as $Z_{500}$ and $T_{850}$. 
Performance is compared against LaDCast models family: small (375M) and big (1.6B).

\begin{table*}[t]
\centering
\caption{\textbf{Comparison to LaDCast.} We report RMSE metrics at \textbf{15-day} forecast horizons. The evaluated variables include atmospheric fields -- UW500 (u-component of wind at 500 hPa), T500 (temperature at 500 hPa), G500 (geopotential at 500 hPa), and SH500 (specific humidity at 500 hPa) -- as well as surface fields -- SLP (sea level pressure), 10m-UW and 10m-VW (u- and v-components of wind at 10 meters), and T2M (temperature at 2 meters).}
\label{tab:mse15}

\begin{tabular}{llccccccccc}
 \toprule
 \textbf{Method} & \textbf{UW500} & \textbf{T500} &
 \textbf{G500} & \textbf{SH500}, $10^{-3}$ & \textbf{SLP} & \textbf{10m-UW} & \textbf{10m-VW}&\textbf{T2M}\\ 
 \hline
 LaDCast, 375M&8.14&3.07&784.21&0.832&690.74&3.81& 3.93 &2.59 \\
 % \hline
 LaDCast, 1.6B &\textbf{8.08}&\textbf{3.04}&\textbf{774.18}&\textbf{0.796}&\textbf{684.29}&\textbf{3.77}& \textbf{3.91} &2.54 \\
 % \hline
Marchuk, 276M&8.15&3.08&785.91&0.803&697.89&3.80& 3.95&\textbf{2.52} \\
 % \hline
Climatology &8.55&3.18&810.57&0.857&711.21&3.96& 4.02&2.66 \\ 
 \bottomrule
\end{tabular}
\end{table*}

\begin{table*}[t]
\centering
\caption{\textbf{Comparison to LaDCast.} We report RMSE metrics at \textbf{30-day} forecast horizons. The evaluated variables include atmospheric fields -- UW500 (u-component of wind at 500 hPa), T500 (temperature at 500 hPa), G500 (geopotential at 500 hPa), and SH500 (specific humidity at 500 hPa) -- as well as surface fields -- SLP (sea level pressure), 10m-UW and 10m-VW (u- and v-components of wind at 10 meters), and T2M (temperature at 2 meters).}
\label{tab:mse30}

\begin{tabular}{llccccccccc}
 \toprule
 \textbf{Method}& \textbf{UW500} & \textbf{T500} &
 \textbf{G500} & \textbf{SH500}, $10^{-3}$ & \textbf{SLP} & \textbf{10m-UW} & \textbf{10m-VW}&\textbf{T2M}\\ 
 \hline
 LaDCast, 375M&8.47&3.26&840.58&0.883&727.40&3.93& 4.01 &2.80 \\
 % \hline
 LaDCast, 1.6B &\textbf{8.40}&3.17&815.53&\textbf{0.825}&\textbf{713.08}&3.87& \textbf{3.97} &2.66 \\ 
 % \hline
Marchuk, 276M &\textbf{8.40}&\textbf{3.16}&\textbf{809.15}&0.836&715.79&\textbf{3.86}& 3.99&\textbf{2.64} \\ 
 % \hline
Climatology &8.55&3.18&810.57&0.857&711.21&3.96& 4.02&2.66 \\ 
 \bottomrule
\end{tabular}
\end{table*}

\begin{figure}[t]
    \centering
    \includegraphics[width=1.0\linewidth]{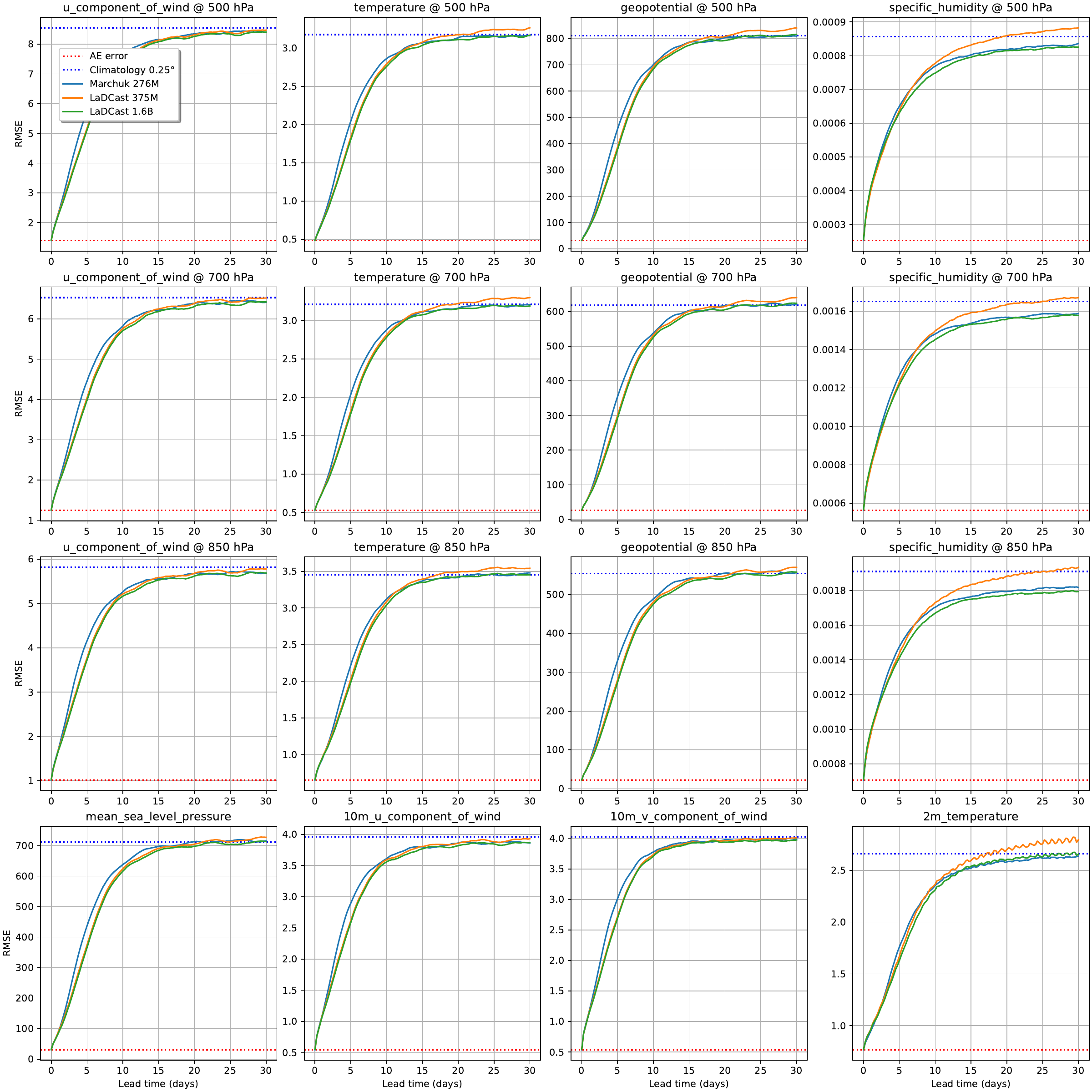}
    \caption{\textbf{RMSE} comparison. We evaluate LaDCast and Marchuk on the WeatherBench-2 benchmark over a 30-day prediction horizon. The atmospheric variables include the u-component of wind, temperature, geopotential, and specific humidity at 500, 750, and 850 hPa levels. Surface variables include mean sea level pressure, 10-meter wind u- and v-components, and temperature at 2 meters. Marchuk outperforms the small LaDCast model and achieves comparable performance to the large LaDCast model. Beyond approximately 30 days, both models’ forecasts converge toward climatology, indicating that extending the prediction horizon further provides limited practical value.
    }
    \label{fig:rmse}
\end{figure}

\paragraph{Evaluation Protocol.} All evaluation metrics presented in the subsequent figures and tables (Figures~\ref{fig:rmse}--\ref{fig:spread} and Tables~\ref{tab:mse15}--\ref{tab:crps30}) were computed using 100 initialization dates sampled uniformly from the test years 2018 and 2021, strictly following the WeatherBench-2 evaluation framework.

\textbf{Executive summary.} The 276M-parameter Marchuk model consistently outperforms the LaDCast 375M baseline across the evaluated metrics and attains performance comparable to the much larger LaDCast 1.6B model. Moreover, the 276M variant of Marchuk achieves an approximately $6\times$ speedup relative to LaDCast 1.6B (see Table~\ref{tab:inference-speed}) while maintaining similar quantitative accuracy.

\textbf{Deterministic metrics.} Deterministic performance is reported in terms of latitude-weighted RMSE and anomaly correlation coefficient (ACC) in Figures~\ref{fig:rmse} and \ref{fig:acc}; these plots also include the DC-AE reconstruction error and climatology baselines for reference. We evaluate forecasts up to a 30-day horizon: for lead times shorter than 30 days, both Marchuk and LaDCast outperform climatology on nearly all considered variables. Tables~\ref{tab:mse15} and \ref{tab:mse30} summarize aggregated error statistics at 15- and 30-day horizons, complementing the figures which display the temporal evolution of forecast error over the entire 30-day period.

\begin{figure}[t]
    \centering
    \includegraphics[width=0.7\linewidth]{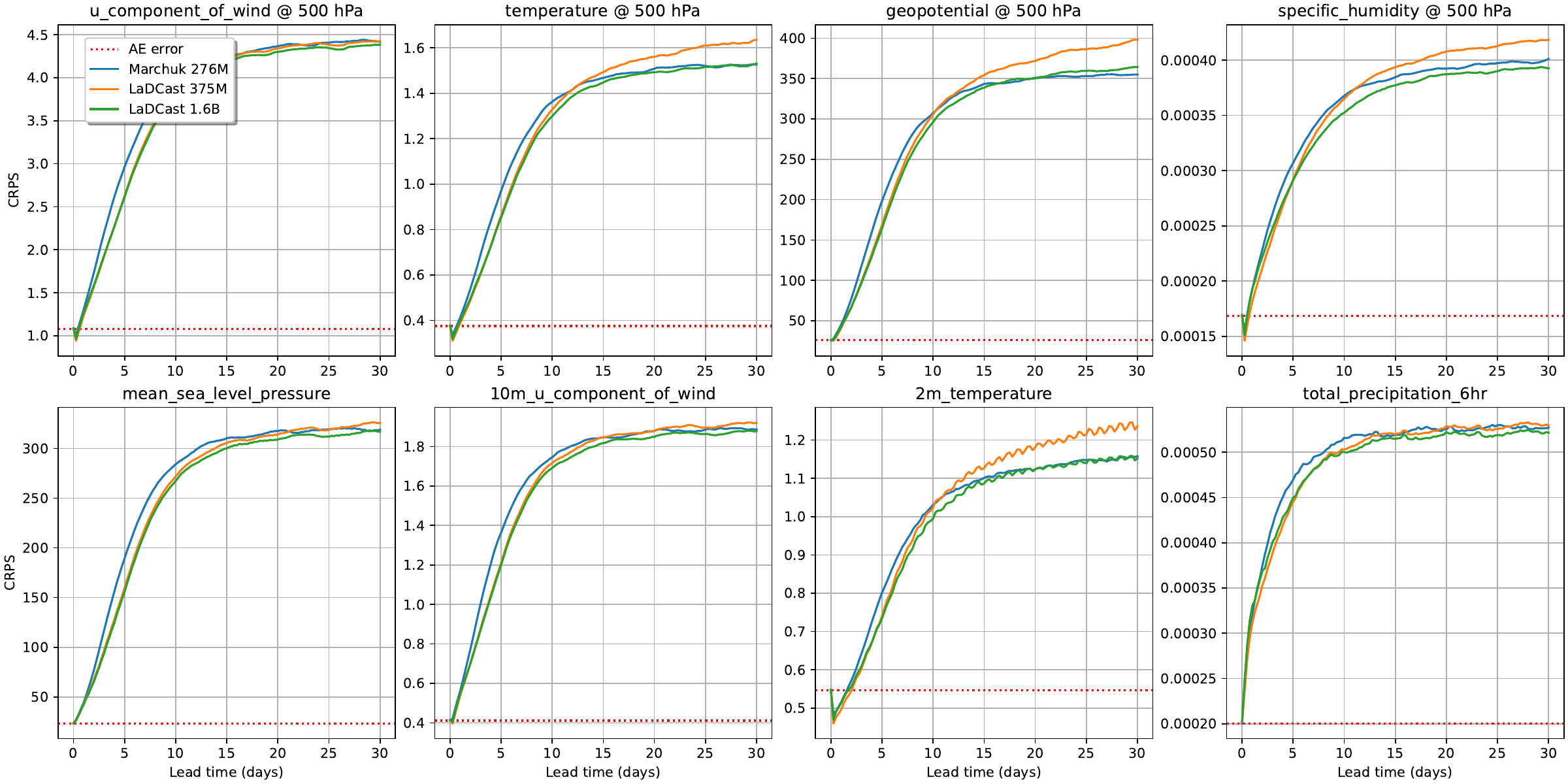}
    \caption{\textbf{CRPS} Ensemble Metrics Comparison. Figure illustrates the evolution of CRPS over a 30-day forecast horizon. The first row presents results for atmospheric variables at the 500 hPa level, including the u-component of wind, temperature, geopotential, and specific humidity. The second row shows CRPS for surface variables, namely mean sea level pressure, the 10-meter u-component of wind, 2-meter temperature, and total accumulated precipitation over 6 hours.
    }
    \label{fig:crps}
\end{figure}

\textbf{Ensemble metrics.} 
Figures~\ref{fig:crps}, \ref{fig:skill} and \ref{fig:spread}, together with Tables~\ref{tab:crps15} ~\ref{tab:crps30}, present probabilistic (ensemble) evaluation metrics that distinguish a generative forecasting system from a deterministic regression baseline. The \emph{skill} metric quantifies the average agreement between the ensemble and the observations, whereas the \emph{spread} measures the ensemble's internal dispersion (pairwise differences between members). The Continuous Ranked Probability Score (CRPS) provides a single summary of probabilistic performance by jointly reflecting agreement (skill) and dispersion (sharpness/spread).

Our results indicate that Marchuk attains higher ensemble skill than the baseline while exhibiting a modestly reduced spread. We attribute this pattern primarily to the larger conditioning window used by Marchuk (one day of context) relative to LaDCast (single-step context, i.e., 6 hours): a longer conditioning interval improves trajectory agreement but can also constrain ensemble variability, yielding lower spread even as average forecast accuracy increases.

\begin{table*}[t]
\centering
\caption{\textbf{Comparison to LaDCast.} We report CRPS metrics at \textbf{15-day} forecast horizons. The evaluated variables include atmospheric fields -- UW500 (u-component of wind at 500 hPa), T500 (temperature at 500 hPa), G500 (geopotential at 500 hPa), and SH500 (specific humidity at 500 hPa) -- and surface fields -- SLP (sea level pressure), 10m-UW (u-component of wind at 10 meters), T2M (temperature at 2 meters), and TP-6h (total accumulated precipitation over the last 6 hours).}
\label{tab:crps15}

\begin{tabular}{llccccccccc}
 \toprule
 \textbf{Method}& \textbf{UW500} & \textbf{T500} &
 \textbf{G500} & \textbf{SH500}, $10^{-3}$ & \textbf{SLP} & \textbf{10m-UW} & \textbf{T2M}&\textbf{TP-6h, $10^{-3}$}\\ 
 \hline
 LaDCast, 375M&4.23&1.49&354.07&0.394&305.53&1.85&1.13&0.520 \\
 % \hline
 LaDCast, 1.6B&\textbf{4.19}&\textbf{1.45}&\textbf{338.43}&\textbf{0.377}&\textbf{299.99}&\textbf{1.82}&\textbf{1.09}&\textbf{0.515} \\
 % \hline
Marchuk, 276M &4.26&1.47&343.34&0.385&310.33&1.85&1.10&0.519\\

 \bottomrule
\end{tabular}
\end{table*}

\begin{table*}[t]
\centering
\caption{\textbf{Comparison to LaDCast.} We report CRPS metrics at \textbf{30-day} forecast horizons. The evaluated variables include atmospheric fields -- UW500 (u-component of wind at 500 hPa), T500 (temperature at 500 hPa), G500 (geopotential at 500 hPa), and SH500 (specific humidity at 500 hPa) -- and surface fields -- SLP (sea level pressure), 10m-UW (u-component of wind at 10 meters), T2M (temperature at 2 meters), and TP-6h (total accumulated precipitation over the last 6 hours).}
\label{tab:crps30}

\begin{tabular}{llccccccccc}
 \toprule
 \textbf{Method}& \textbf{UW500} & \textbf{T500} &
 \textbf{G500} & \textbf{SH500}, $10^{-3}$ & \textbf{SLP} & \textbf{10m-UW} & \textbf{T2M}&\textbf{TP-6h, $10^{-3}$}\\ 
 \hline
LaDCast, 375M&4.43&1.64&398.00&0.419&325.58&1.92&1.24&0.530 \\
 % \hline
LaDCast, 1.6B&\textbf{4.38}&\textbf{1.53}&364.10&\textbf{0.393}&\textbf{316.72}&\textbf{1.88}&\textbf{1.15}&\textbf{0.522} \\ 
 % \hline
Marchuk, 276M&4.42&\textbf{1.53}&\textbf{354.93}&0.401&318.99&1.89&1.16&0.527\\
 \bottomrule
\end{tabular}
\end{table*}

\subsection{Extreme Event Case Study: The January 2026 Moscow Cold Wave}

% To evaluate extreme-event forecasting ability, we conduct a case study of the \textbf{end January 2026 Moscow frost}. During this event, nighttime temperatures dropped below $-20^\circ$C.
To evaluate the model's capacity to predict extreme out-of-distribution events, we conducted a case study of the severe cold wave that struck Moscow in late January 2026. During this event, nighttime temperatures plummeted below $-20^\circ$C. For this experiment, we defined the target region using a bounding box over Moscow (55–58° N, 35–39° E) and generated predictive ensembles of 250 members.

\begin{figure}[t]
    \centering
    \includegraphics[width=0.5\linewidth]{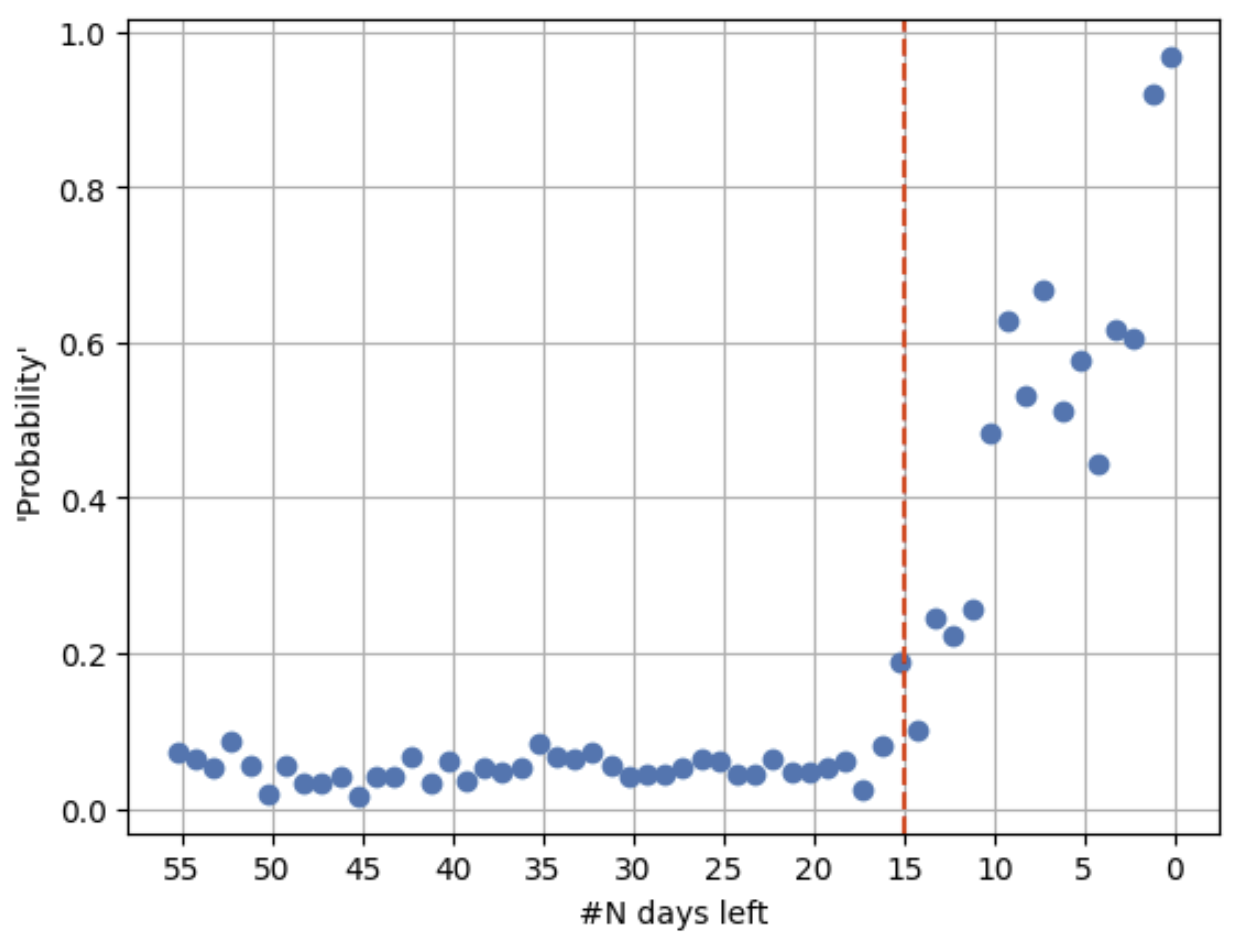}
    \caption{
    \textbf{Extreme-Event Forecasting Ability.} We evaluate the model’s capability to predict extreme events by analyzing the probability of a cold-weather event preceding the Moscow frost day (midnight, January 26, 2026). Specifically, we plot the probability as a function of lead time, defined as the number of days before the event. The probability is computed as the fraction of ensemble members predicting temperatures within the range of $-22^\circ C \pm 3^\circ C$. This analysis provides insight into the model’s ability to capture the likelihood of extreme temperature events over extended forecast horizons.
    }
    \label{fig:moscow-frost}
\end{figure}

The results in Figure~\ref{fig:moscow-frost} demonstrate that at lead times greater than 15 days, the model predicts only baseline climatological probabilities ($\sim$5--10\%). However, as the event approaches, Marchuk predicts increasing anomaly probability reaching approximately \textbf{80--90\%} close to the event date.

\subsection{Ablations}

We performed a set of controlled ablation experiments to quantify the influence of several architectural and training choices on Marchuk's forecasting performance. Specifically, we evaluated (i) positional encoding schemes including trainable spatial embeddings, (ii) fixed input window lengths, (iii) temporal skip intervals between latent frames, (iv) training-horizon strategies (variable vs.\ fixed horizons), (v) explicit timestamp conditioning via cross-attention, and (vi) reductions of the initial context length at inference.

For all ablation experiments, we report RMSE metrics computed on eight selected weather maps at the 15-day forecast horizon. The atmospheric maps include the u-component of wind, temperature, geopotential, and specific humidity at the 500 hPa level. The surface maps consist of sea level pressure, the 10-meter u- and v-components of wind, and temperature at 2 meters.

\paragraph{Positional encodings (RoPE, GeoRoPE, and learned embeddings).}
We evaluated the impact of different positional-encoding schemes on forecast skill by comparing three configurations: (i) full 3-D RoPE (rotary positional embeddings applied over both spatial and temporal dimensions), (ii) GeoRoPE as proposed in the LaDCast paper (a geometry-aware decomposition into latitude and longitude components), and (iii) a hybrid scheme combining 1-D RoPE along the temporal axis with learnable 2-D spatial embeddings. Replacing GeoRoPE with the hybrid, trainable-embedding scheme produces consistent improvements in RMSE metrics. The results for the 15 day RMSE horizon are reported in Table~\ref{tab:rope-ablation}.

\begin{table}[h]
\centering
\caption{Evaluation of Marchuk with different positional encoding schemes: RoPE3D, GeoRoPE, and RoPE 1D with trainable spatial embeddings. The hybrid configuration achieves the best performance in terms of RMSE over a 15-day forecast horizon.}
\label{tab:rope-ablation}
\begin{tabular}{lcccccccc}
 \toprule
 \textbf{Config} & \textbf{UW500} & \textbf{T500} &
 \textbf{G500} & \textbf{SH500}, $10^{-3}$ & \textbf{SLP} & \textbf{10m-UW} & \textbf{10m-VW} &\textbf{T2M}\\ 
 \hline
 RoPE 3D&8.63&3.23&829.83&0.851&730.81&3.98&4.11&2.66\\
 % \hline
 GeoRoPE&8.53&3.20&821.28&0.848&729.82&3.97&4.09&2.67\\
 % \hline
RoPE 1D + trainable 2D&\textbf{8.42}&\textbf{3.18}&\textbf{808.11}&\textbf{0.841}&\textbf{719.79}&\textbf{3.94}&\textbf{4.05}&\textbf{2.65}\\
 \bottomrule
\end{tabular}
\end{table}

\textbf{Training with Fixed Window.} We train Marchuk using varying context sizes, corresponding to fixed video durations of 6, 12, 24, and 48 hours, which map to 1, 2, 4, and 8 global weather frames, respectively. As the context size increases, we observe a consistent improvement in quantitative performance metrics, indicating that longer temporal context enhances the model’s predictive capability Table~\ref{tab:window-ablation}.

\begin{table}[h]
\centering
\caption{Effect of fixed-window training on model performance. Increasing the context window length consistently improves RMSE across all evaluated variables.}
\label{tab:window-ablation}
\begin{tabular}{lcccccccc}
 \toprule
 \textbf{Config} & \textbf{UW500} & \textbf{T500} &
 \textbf{G500} & \textbf{SH500}, $10^{-3}$ & \textbf{SLP} & \textbf{10m-UW} & \textbf{10m-VW} &\textbf{T2M}\\ 
 \hline
 6 hours&8.61&3.21&820.05&0.864&727.83&4.02&4.10&2.68\\
 % \hline
 12 hours&8.54&3.15&806.22&0.842&714.67&\textbf{3.93}&\textbf{4.08}&2.63\\
 % \hline
24 hours&8.51&3.17&809.98&0.834&715.94&3.95&\textbf{4.08}&2.66\\
 % \hline
48 hours&\textbf{8.39}&\textbf{3.14}&\textbf{804.05}&\textbf{0.828}&\textbf{713.98}&\textbf{3.93}&\textbf{4.08}&\textbf{2.61}\\
 \bottomrule
\end{tabular}
\end{table}

\textbf{Skip Interval: 1 Hour vs. 6 Hours vs. 12 Hours.} We conduct an ablation study on the temporal skip interval used during training and inference. Given that the ERA5 dataset provides data at an hourly resolution, we vary the skip interval to reduce the number of autoregressive steps required for long-term forecasting. Specifically, we evaluate intervals of 1, 6, and 12 hours. Our results show that a 6-hour skip interval achieves the best performance for forecasts extending up to 15 days, providing a favorable trade-off between predictive accuracy and the number of autoregressive steps Table~\ref{tab:skip-interval-ablation}.

\begin{table}[h]
\centering
\caption{Effect of temporal skip interval on model performance. Training with different timestep intervals (1, 6, and 12 hours) shows that a 6-hour interval achieves the best RMSE performance, balancing temporal resolution and autoregressive efficiency.}
\label{tab:skip-interval-ablation}
\begin{tabular}{lcccccccc}
 \toprule
 \textbf{Config} & \textbf{UW500} & \textbf{T500} &
 \textbf{G500} & \textbf{SH500}, $10^{-3}$ & \textbf{SLP} & \textbf{10m-UW} & \textbf{10m-VW} &\textbf{T2M}\\ 
 \hline
 1 hour&8.60&3.20&829.13&0.847&733.35&3.98&4.10&2.72\\
 % \hline
 6 hours&\textbf{8.49}&\textbf{3.14}&\textbf{814.46}&\textbf{0.834}&\textbf{717.28}&\textbf{3.94}&\textbf{4.05}&\textbf{2.65}\\
 % \hline
 12 hours&8.65&3.32&851.86&0.858&744.54&4.02&4.21&2.75\\
 \bottomrule
\end{tabular}
\end{table}

\textbf{Inference Strategy.} We evaluate the model under varying prediction horizons at inference time while employing a dynamic sequence length during training. To determine the optimal context size, we conduct four experiments with configurations spanning from 1-day to 8-day input sequences. This approach enables the model to adapt to different temporal contexts and facilitates a systematic analysis of how input length influences forecasting performance Table~\ref{tab:train-strategy-abl}.

\begin{table}[h]
\centering
\caption{Impact of inference configuration on model performance. Varying the input context length from 1 to 8 days shows that intermediate horizons (2-4 days) yield the best RMSE performance.}
\label{tab:train-strategy-abl}
\begin{tabular}{lcccccccc}
 \toprule
 \textbf{Config} & \textbf{UW500} & \textbf{T500} &
 \textbf{G500} & \textbf{SH500}, $10^{-3}$ & \textbf{SLP} & \textbf{10m-UW} & \textbf{10m-VW} &\textbf{T2M}\\ 
 \hline
 1 day&8.43&3.15&808.56&0.827&717.27&3.92&4.06&2.68\\
 % \hline
 2 days&8.34&\textbf{3.13}&\textbf{800.60}&\textbf{0.816}&\textbf{715.86}&3.92&\textbf{4.04}&2.59\\
 % \hline
 4 days&\textbf{8.31}&3.15&815.62&\textbf{0.816}&728.41&\textbf{3.89}&4.08&\textbf{2.57}\\
% \hline
 8 days&8.53&3.20&826.77&0.817&741.44&3.93&4.11&2.65\\
 \bottomrule
 % + tune 16 days&8.61&3.25&837.33&0.822&737.13&3.95&4.10&2.69\\
 % \hline
\end{tabular}
\end{table}

\textbf{Adding Timestamp in Cross-Attention.} We investigate the effect of incorporating timestamp information as a conditioning signal in the cross-attention mechanism. Our experiments compare models trained with and without timestamp conditioning. The results demonstrate that including timestamp consistently improves performance, yielding lower RMSE across all evaluated parameters Table ~\ref{tab:cond-ablation}.

\begin{table}[h]
\centering
\caption{Ablation study on the effect of including timestamp information in Cross-Attention. Incorporating timestamp as a conditioning signal improves RMSE across all evaluated variables.}
\label{tab:cond-ablation}
\begin{tabular}{lcccccccc}
 \toprule
 \textbf{Config} & \textbf{UW500} & \textbf{T500} &
 \textbf{G500} & \textbf{SH500}, $10^{-3}$ & \textbf{SLP} & \textbf{10m-UW} & \textbf{10m-VW} &\textbf{T2M}\\ 
 \hline
 w/o timestamp&\textbf{8.39}&\textbf{3.14}&804.05&0.828&713.98&3.93&4.08&2.61\\
 % \hline
 w timestamp &\textbf{8.39}&\textbf{3.14}&\textbf{799.18}&\textbf{0.827}&\textbf{708.66}&\textbf{3.92}&\textbf{4.05}&\textbf{2.59}\\
 \bottomrule
\end{tabular}
\end{table}

\textbf{Initial Context Length.} We evaluate the impact of reducing the input context length on model performance. Decreasing the context to 4 points (corresponding to 1 day of weather maps) does not degrade predictive accuracy. To optimize computational efficiency while maintaining acceptable metric values, we adopt this reduced context length. Predictions are generated autoregressively by sampling the next 4 days, selecting the fourth day, and rolling out subsequent forecasts in a similar manner Table~\ref{tab:context-ablation}.
\begin{table}[h]
\caption{Ablation study on input context length. Reducing the context to 4 points (1 day of weather maps) maintains predictive accuracy while improving computational efficiency.}
\label{tab:context-ablation}
\centering
\begin{tabular}{lcccccccc}
 \toprule
 \textbf{Config} & \textbf{UW500} & \textbf{T500} &
 \textbf{G500} & \textbf{SH500}, $10^{-3}$ & \textbf{SLP} & \textbf{10m-UW} & \textbf{10m-VW} &\textbf{T2M}\\ 
 \hline
 12 hours&8.50&3.17&822.71&0.838&739.36&4.00&4.12&\textbf{2.62}\\
 % \hline
 1 day&\textbf{8.42}&\textbf{3.18}&\textbf{811.02}&0.839&727.00&\textbf{3.96}&\textbf{4.09}&2.65\\
 % \hline
 2 days&8.51&3.20&817.01&\textbf{0.835}&\textbf{715.08}&\textbf{3.96}&\textbf{4.09}&2.65\\
 \bottomrule
\end{tabular}
\end{table}

%% file: sections/5_conclusion.tex
\section{Conclusion}
\label{sec:conclusion}

We have introduced Marchuk, a latent flow-matching model for probabilistic global weather forecasting on mid-range to sub-seasonal timescales (up to 30 days). Marchuk leverages a highly compressed latent representation (64× spatial compression) and flow-matching, achieving an excellent balance between accuracy and efficiency. In experiments on ERA5-based WeatherBench data, Marchuk (276M parameters) matches or exceeds the performance of much larger models (e.g. LaDCast at 1.6B params) while running an order of magnitude faster. Notably, Marchuk yields comparable lat-weighted RMSE and ACC, and competitive CRPS/Spread scores, demonstrating that compact latent transformers can capture the essential atmospheric dynamics. We also show that Marchuk’s ensembles accurately characterize uncertainty, as evidenced by case studies of extreme events (e.g. the January 2026 Moscow frost) where the probabilistic forecasts successfully identified the impending anomaly.

The ablation studies inform our design choices. We find that learnable 2D spatial embeddings combined with temporal RoPE outperform hand-coded GeoRoPE, and that a moderate 1-day context window (4 six-hour frames) strikes a good trade-off between accuracy and compute. Training with a variable horizon of 1-8 day horizons produces robust multi-week forecasts. These insights help us select the final Marchuk configuration used in the main experiments.

Despite these successes, limitations remain. The primary bottleneck is the DC-AE’s reconstruction error, which inherently bounds forecast fidelity. Like LaDCast, Marchuk is trained on reanalysis data, so its deployment to real-time forecasting would require integrating data-assimilation or multi-encoder schemes. The spatial resolution (1.5°) also constrains fine-scale features such as tropical cyclone intensity. Future work should explore higher-resolution latent compressions, physics-informed losses, and assimilation of live observations to further improve skill.

%% file: sections/6_determine.tex
\begin{figure}[t]
    \centering
    \includegraphics[width=1.0\linewidth]{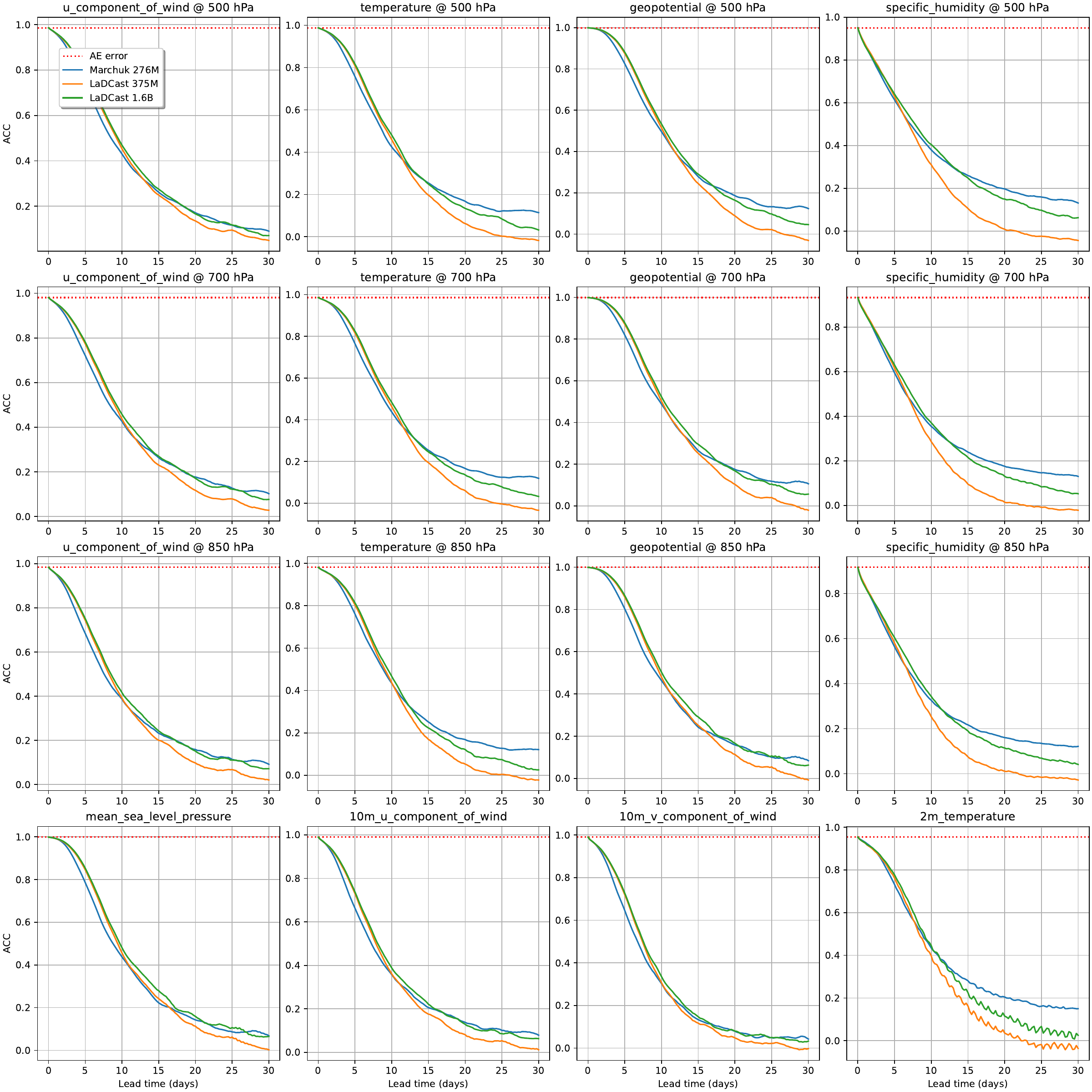}
    \caption{\textbf{ACC} score. We evaluate model performance using ACC across several weather variables. The results, illustrated in the figure, show that our model exhibits the smallest decline in ACC over time compared to both LaDCast models.
    }
    \label{fig:acc}
\end{figure}
\label{sec:determine}

%% file: sections/7_probabalistic.tex
\begin{figure}[t]
    \centering
    \includegraphics[width=1.0\linewidth]{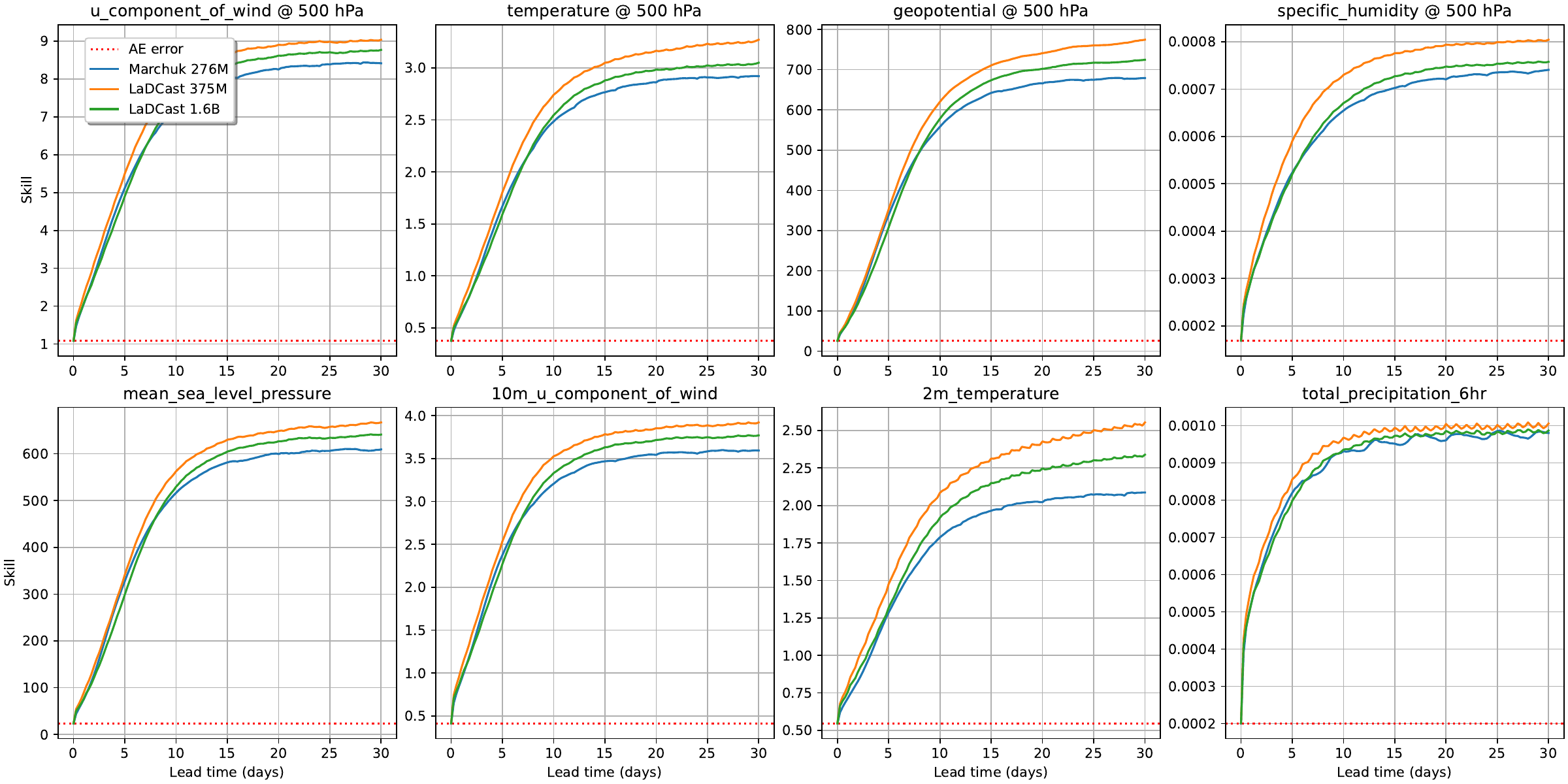}
    \caption{The \textbf{CRPS Skill} score evaluates how closely ensemble predictions align with the observed outcomes on average. Our model outperforms both the small and large LaDCast variants, demonstrating improved probabilistic forecasting accuracy.
    }
    \label{fig:skill}
\end{figure}

\begin{figure}[t]
    \centering
    \includegraphics[width=1.0\linewidth]{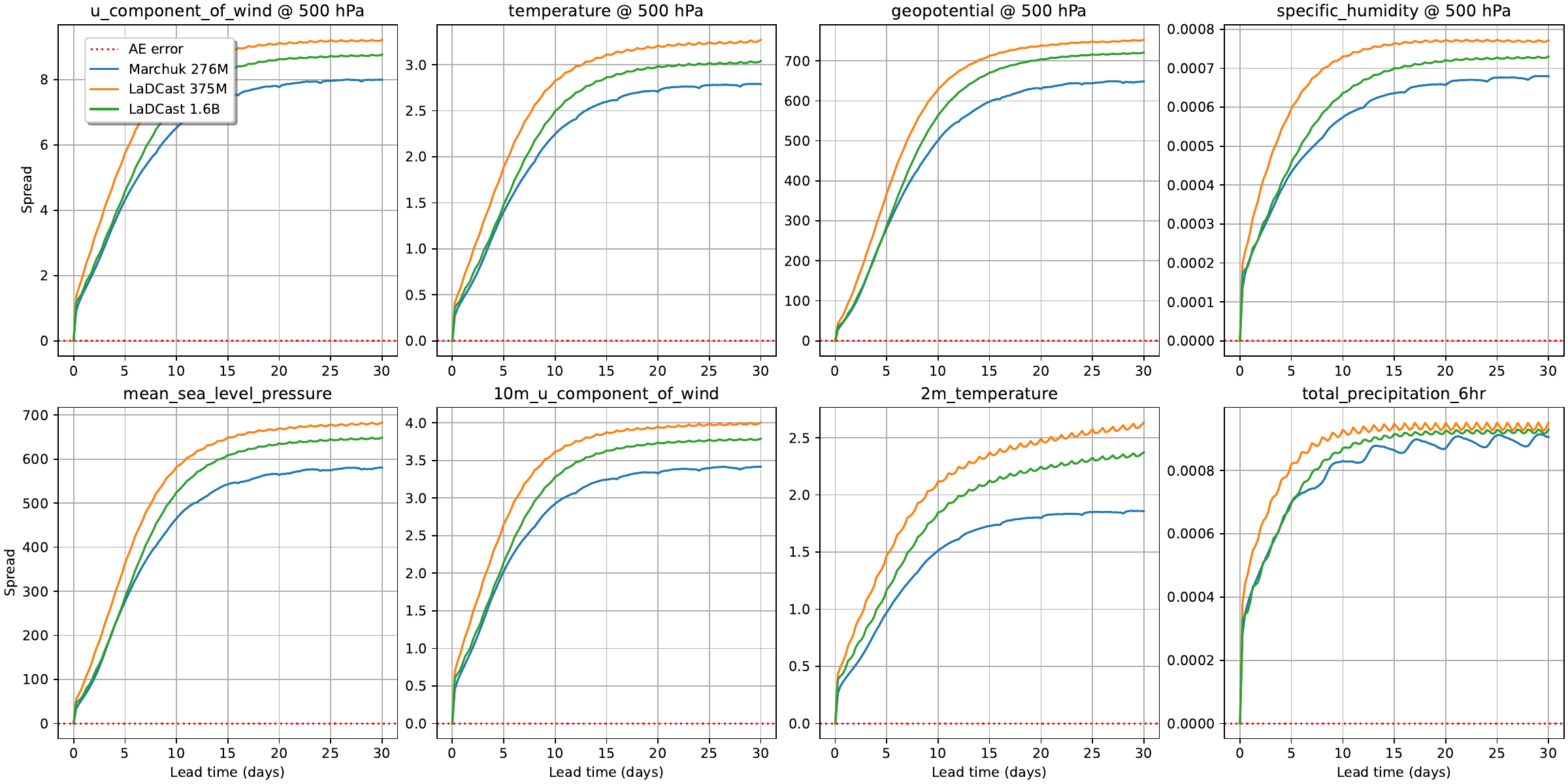}
    \caption{\textbf{CRPS Spread.} The CRPS spread reflects the diversity among ensemble members by measuring pairwise differences within the ensemble. Our model exhibits lower spread compared to LaDCast, indicating reduced ensemble diversity. This effect can be attributed to stronger conditioning: while LaDCast conditions on a single weather map, Marchuk conditions on the four most recent maps, leading to more constrained and less diverse predictions.
    }
    \label{fig:spread}
\end{figure}